# Alternating Projections for Learning with Expectation Constraints

**Kedar Bellare**
Dept. of Computer Science
University of Massachusetts
Amherst, MA 01003
kedarb@cs.umass.edu

**Gregory Druck**
Dept. of Computer Science
University of Massachusetts
Amherst, MA 01003
gdruck@cs.umass.edu

**Andrew McCallum**
Dept. of Computer Science
University of Massachusetts
Amherst, MA 01003
mccallum@cs.umass.edu

## Abstract

We present an objective function for learning with unlabeled data that utilizes auxiliary expectation constraints. We optimize this objective function using a procedure that alternates between information and moment projections. Our method provides an alternate interpretation of the *posterior regularization framework* (Graca et al., 2008), maintains uncertainty during optimization unlike *constraint-driven learning* (Chang et al., 2007), and is more efficient than *generalized expectation criteria* (Mann & McCallum, 2008). Applications of this framework include minimally supervised learning, semi-supervised learning, and learning with constraints that are more expressive than the underlying model. In experiments, we demonstrate comparable accuracy to *generalized expectation criteria* for minimally supervised learning, and use expressive structural constraints to guide semi-supervised learning, providing a 3%-6% improvement over state-of-the-art *constraint-driven learning*.

## 1 Introduction

Maximum entropy parameter estimation, in which the objective is to choose the most uncertain distribution that satisfies a set of expectation constraints, is widely used for supervised learning (Berger et al., 1996; Lafferty et al., 2001). Recently there has been interest in learning with unlabeled data using auxiliary expectation constraints. For example, limited prior knowledge about the labels that words are likely to indicate has been used to learn accurate discriminative models for NLP tasks (Druck et al., 2008; Mann & McCallum, 2008), and structural constraints on output variables have been used to guide semi-supervised learning (Graca et al., 2008; Chang et al., 2007).

Several frameworks have been proposed for learning with auxiliary constraints. *Generalized expectation* (GE) (Druck et al., 2008; Mann & McCallum, 2008) enforces arbitrary expectation constraints on a underlying Markov random field model by penalizing constraint violation in the objective function. *EM with posterior constraints* (EMPC) (Graca et al., 2008) (also called *posterior regularizaton* more recently (Ganchev et al., 2009)) modifies the E-step of the EM algorithm to project the model posterior onto the set of distributions that satisfy the auxiliary constraints. *Constraint-driven learning* (CODL) (Chang et al., 2007) re-estimates model parameters using the $n$-best outputs according to the model and auxiliary constraint violation penalties. Importantly, these frameworks learn with auxiliary constraints that are not in one-to-one correspondence with the parameterization of the desired probabilistic model. This flexibility can be used to learn a feature-rich model with only a few auxiliary constraints, or specify auxiliary constraints that would make inference in the underlying model intractable.

In this paper we present an objective function for learning with auxiliary expectation constraints. We optimize this objective function using *alternating projections* (AP). In the first step of optimization, we find the auxiliary distribution that is the information projection of the model distribution onto the set of distributions that satisfy the auxiliary expectation constraints. In the next step, we re-estimate the model parameters using a moment projection that matches the moments of the model and auxiliary distribution. This method provides an alternate interpretation of EMPC, as we discuss in Section 4.2. Unlike CODL, which uses a point estimate of labels for unlabeled data, AP preserves uncertainty between projection steps. AP also provides a more efficient method for learning with constraints in structured output models than GE, and can additionally be used in an online setting.

In experiments, we compare with GE for minimally supervised learning of logistic regression models for document classification, and conditional random field models for sequence labeling. The experiments show that AP gives comparable accuracy while providing computational advantages. Additionally we use expressive structural constraints to guide semi-supervised learning of sequence labeling models, outperforming previously state-of-the-art CODL by 3-6% on a standard reference extraction task.

## 2 Approach

Let $x \in \mathcal{X}$ and $y \in \mathcal{Y}$ be the input and output random variables respectively. We assume that input variables $x$ are drawn from an underlying unknown distribution $\hat{p}(x)$. Given an input $x$, the output $y$ is generated according to the unknown conditional distribution $\hat{p}(y|x)$. Our goal during learning is to estimate this conditional distribution. Let $\mathbf{f} = (f_1, f_2, \ldots, f_l)$ be a set of sufficient statistics or *features* where $f_i : \mathcal{X} \times \mathcal{Y} \to \mathbb{R}, \forall i = 1, \ldots, l$.

### 2.1 Maximum Entropy Estimation

Let $\mathcal{D} = \{(x_1, y_1), (x_2, y_2), \ldots, (x_m, y_m)\}$ be a data set of labeled examples. The principle of maximum entropy (Berger et al., 1996) states that of all distributions $q(y|x)$ that match the moments of the data sample, we should choose the one with the maximum entropy. This results in the following constrained optimization problem,

$$\begin{aligned}
\max_q \quad & \sum_{i=1}^m H[q(y|x_i)] \\
\text{s.t.} \quad & \sum_{i=1}^m E_{q(y|x_i)}[\mathbf{f}(x_i, y)] = \sum_{i=1}^m \mathbf{f}(x_i, y_i) \\
& \sum_y q(y|x_i) = 1, \forall i = 1 \ldots m \\
& q(y|x_i) \geq 0, \forall y, i = 1 \ldots m,
\end{aligned} \quad (1)$$

where $H(\cdot)$ is the entropy of the distribution. The dual of this optimization problem is equivalent to maximizing the log-likelihood of the data $\mathcal{L}(\lambda; \mathcal{D}) = \sum_{i=1}^m \log(q_\lambda(y_i|x_i))$ for the exponential distribution $q_\lambda(y|x) = \frac{1}{Z_\lambda(x)} \exp(\lambda \cdot \mathbf{f}(x, y))$ (Berger et al., 1996). The model parameters $\lambda$ correspond to the Lagrange multipliers in the unconstrained primal objective for Equation (1). Here $Z_\lambda(x) = \sum_{y' \in \mathcal{Y}} \exp(\lambda \cdot \mathbf{f}(x, y'))$ is the partition function.

The objective defined above can be generalized to include other convex constraints (or penalties) and a divergence from a base distribution (Dudik, 2007). Given a base distribution $q_0(y|x)$ and a convex function $U(\sum_{i=1}^m E_{q(y|x_i)}[\mathbf{f}(x_i, y)])$ we can optimize the primal problem,

$$\min_q \sum_{i=1}^m D[q(y|x_i) \| q_0(y|x_i)] + U(\sum_{i=1}^m E_{q(y|x_i)}[\mathbf{f}(x_i, y)]), \quad (2)$$

where $D(q(y|x) \| q_0(y|x)) = \sum_y q(y|x) \log\left(\frac{q(y|x)}{q_0(y|x)}\right)$ is the KL-divergence between conditional distributions $q$ and $q_0$. Although we use KL-divergence in this paper, other divergence measures can be used as well (Altun & Smola, 2006). Using Fenchel's duality theorem (Dudik, 2007), the dual of Equation (2) is,

$$\max_\lambda \sum_{i=1}^m -\log[\sum_y q_0(y|x_i) \exp(\lambda \cdot \mathbf{f}(x_i, y))] - U^*(-\lambda), \quad (3)$$

where $U^*$ is the conjugate of the convex function $U$ and $\lambda \in \mathbb{R}^l$ are the dual parameters. The primal variables $q(y|x)$ with respect to dual parameters $\lambda$ are $q(y|x) \propto q_0(y|x) \exp(\lambda \cdot \mathbf{f}(x, y))$. Equation (1) uses a uniform base distribution $q_0(y|x)$ and equality constraints as the convex function $U$. Similarly, when the convex function $U$ is the $\mathbb{L}_2$ penalty $\frac{1}{2\alpha} \| \sum_{i=1}^m \mathbf{f}(x_i, y_i) - E_{q(y|x_i)}[\mathbf{f}(x_i, y)] \|_2^2$ and the base distribution $q_0(y|x)$ is uniform, we obtain a minimization of the $\mathbb{L}_2$-regularized log-loss over the data $\mathcal{D}$,

$$\mathcal{L}(\lambda; \mathcal{D}) = \min_\lambda \sum_{i=1}^m -\log(q_\lambda(y_i|x_i)) + \frac{\alpha}{2} \|\lambda\|_2^2, \quad (4)$$

where $\alpha$ is the weight for the $\mathbb{L}_2$-regularization term and $q_\lambda(y|x) = \frac{1}{Z_\lambda(x)} \exp(\lambda \cdot \mathbf{f}(x, y))$ is an exponential family distribution with $Z_\lambda(x) = \sum_{y'} \exp(\lambda \cdot \mathbf{f}(x, y'))$.

### 2.2 Learning with Unlabeled Data using Auxiliary Constraints

Although maximum entropy is widely used for supervised learning (Berger et al., 1996; Lafferty et al., 2001), it cannot be applied to unsupervised and semi-supervised learning. In the absence of constraints, the maximum entropy distribution on unlabeled data is the uniform distribution. Instead, prior work on semi-supervised learning has focused on minimizing the label entropy on the unlabeled data points (Grandvalet & Bengio, 2004; Jiao et al., 2006). This can be harmful in practice because assigning all instances to a single label minimizes entropy.

Recent work in semi-supervised learning (Chang et al., 2007; Mann & McCallum, 2008) has focused on incorporating prior knowledge into learning. In particular, we may have expectation constraints on auxiliary features $\mathbf{f}' = (f_1', \ldots, f_t')$ that must be satisfied on an unlabeled sample $\mathcal{D}' = \{x_{m+1}, x_{m+2}, \ldots, x_n\}$. Generalized expectation (GE) (Mann & McCallum, 2008) adds additional terms to the objective function that score the model expectations of $\mathbf{f}'$. Published work has focused on minimizing the divergence of model expectations of $\mathbf{f}'$ and user-specified target values $\boldsymbol{u} = (u_1, \ldots, u_t)$ in addition to minimizing the regularized log-loss over labeled data $\mathcal{D}$,

$$\mathcal{L}_{GE}(\lambda) = \min_\lambda \sum_{i=1}^m -\log(p_\lambda(y_i|x_i)) + \frac{\alpha}{2}\|\lambda\|_2^2$$
$$+ \gamma U\Big(\sum_{j=m+1}^n E_{p_\lambda(y|x_j)}[\mathbf{f}'(x_j,y)]\Big), \quad (5)$$

where $\gamma$ is the weight of the GE terms, $p_\lambda(y|x)$ is the model distribution and $U(\cdot)$ is a divergence (e.g. KL or squared divergence) from targets $\boldsymbol{u}$. This objective, however, is expensive to optimize since it requires computing the covariance between auxiliary features $\mathbf{f}'$ and model features $\mathbf{f}$ (cf. Section 4.4). Henceforth, let $\sum_i = \sum_{i=1}^m$ and $\sum_j = \sum_{j=m+1}^n$ be the summations over labeled and unlabeled examples respectively.

Instead of directly optimizing the objective above, we introduce an auxiliary distribution $q(y|x)$ that satisfies general convex constraints $U(\sum_j E_q[\mathbf{f}'(x_j,y)])$ and additionally has low divergence with the model distribution $p_\lambda(y|x)$. The general class of convex constraints $U(\cdot)$ includes the divergence functions used by GE. Thus, we optimize the joint objective,

$$\mathcal{O}(\lambda, q) = \min_{\lambda,q} \sum_i -\log(p_\lambda(y_i|x_i)) + \frac{\alpha}{2}\|\lambda\|_2^2$$
$$+ \gamma\Big[\sum_j D(q(y|x_j)\|p_\lambda(y|x_j))$$
$$+ U(\sum_j E_q[\mathbf{f}'(x_j,y)])\Big], \quad (6)$$

where $\gamma$ weights the relative contribution of the labeled and unlabeled terms in the objective. The same objective can be used in the absence of labeled data by dropping the log-loss term from the objective. The optimization over $q(y|x)$ is similar to the generalized maximum entropy problem described earlier. The dual of this objective with respect to $q(y|x)$ is,

$$\mathcal{O}(\lambda, \mu) = \min_\lambda \max_\mu \sum_i -\log(p_\lambda(y_i|x_i)) + \frac{\alpha}{2}\|\lambda\|_2^2$$
$$-\gamma\Big[\sum_j (\log Z_{\lambda,\mu}(x_j) - \log Z_\lambda(x_j))$$
$$- U^*(-\mu)\Big], \quad (7)$$

where $Z_{\lambda,\mu}(x) = \sum_{y' \in \mathcal{Y}} \exp(\lambda \cdot \mathbf{f}(y',x) + \mu \cdot \mathbf{f}'(y',x))$ is the normalization of the parameterized auxiliary distribution $q_{\lambda,\mu}(y|x) = \frac{1}{Z_{\lambda,\mu}(x)} \exp(\lambda \cdot \mathbf{f}(y,x) + \mu \cdot \mathbf{f}'(y,x))$ and $\mu$ are the parameters of the dual objective.

The convex function $U$ can take many forms (Dudik, 2007). For example:

- $\mathbb{L}_2$ penalty $\frac{1}{2\beta}\|u - \sum_j E_q[f'(x_j,y)]\|_2^2$ with conjugate $U^*(-\mu) = -\mu \cdot u + \frac{\beta}{2}\|\mu\|_2^2$. For example, this can be used in a document classification task ibm vs. mac to indicate that about 95% of documents containing the term "windows" should have class ibm. In this case, the target $u = 0.95N$ where $N$ denotes the number of documents with the term "windows" and the feature functions $f'(x,y) = \mathbb{I}(\text{"windows"} \in x \wedge y = \text{ibm})$ where $\mathbb{I}$ is the indicator function.

- $\mathbb{L}_1$ box constraints $|u - \sum_j E_q[f'(x_j,y)]| \leq \beta$ with conjugate $U^*(-\mu) = -\mu \cdot u + \beta|\mu|$. This constraint can be used to indicate that an auxiliary expectation lies in the interval $[u - \beta, u + \beta]$. Hence, if we know that 70-90% of the $N$ documents containing the term "firewire" are about mac, then our $\mathbb{L}_1$ box constraint has $u = 0.8N$, $\beta = 0.1N$ and $f'(x,y) = \mathbb{I}(\text{"firewire"} \in x \wedge y = \text{mac})$.

- Affine constraints $E_q[f'(x,y)] \leq u$ (Graca et al., 2008) with conjugate $U^*(-\mu) = -\mu \cdot u, \mu \geq 0$. We can use this constraint in a citation extraction task to express a structural constraint that the journal field appears at most once in the sequence. Here, $E_q[f'(x,y)] \leq 1$ with $f'(x,y) = \sum_{i=1}^L \mathbb{I}(y_{i-1} \neq \text{journal}, y_i = \text{journal})$ for label sequence $y = (y_1, y_2, \ldots, y_L)$.

Such constraints serve as substitutes for labels on the unlabeled data and can be expressed locally for each instance or globally on the entire data set.

### 2.3 Learning with Alternating Projections

We optimize the min-max objective (Equation (7)) using coordinate descent on $\lambda$ and $\mu$ parameters, fixing one while optimizing the other. Starting with initial parameters $\lambda^{(0)}, \mu^{(0)}$ we alternate between the two following projection steps ($t = 1, \ldots, T$):

- **I-projection** (information projection): Find parameters $\mu^{(t)}$ such that auxiliary distribution $q$ is close to $p_{\lambda^{(t-1)}}$ and minimizes the convex function $U(\sum_j E_q(\mathbf{f}'(x_j,y)))$. For ease of notation, we assume that the constraints $U$ are $\mathbb{L}_2$ penalty terms $\frac{1}{2\beta}\|\boldsymbol{u} - \sum_j E_q(\mathbf{f}'(x_j,y))\|_2^2$. Hence, $\mu^{(t)}$ is

$$\arg\max_\mu \mu \cdot \boldsymbol{u} - \sum_j \log Z_{\lambda^{(t-1)},\mu}(x_j) - \frac{\beta}{2}\|\mu\|_2^2, \quad (8)$$

where $\beta$ is the regularization constant. The gradient of the above objective is $u - \sum_j E_{q_{\lambda^{(t-1)},\mu}}[\mathbf{f}'(x_j, y)] - \beta\mu$. This gradient matches the auxiliary moments of features $\mathbf{f}'$ to the user-specified targets $u$.

- **M-projection** (moment projection): We now fix the $q$ distribution and find parameters $\lambda^{(t)}$ as,

$$\arg\max_\lambda \sum_i \left[\lambda \cdot \mathbf{f}(x_i, y_i) - \log Z_\lambda(x_i)\right] - \frac{\alpha}{2}\|\lambda\|_2^2$$
$$+ \gamma \sum_j \left[\lambda \cdot E_q[\mathbf{f}(x_j, y)] - \log Z_\lambda(x_j)\right]. \quad (9)$$

The gradient of this equation is given by $\sum_i (\mathbf{f}(x_i, y_i) - E_{p_\lambda}[\mathbf{f}(x_i, y)]) - \alpha\lambda + \gamma \sum_j (E_q[\mathbf{f}(x_j, y)] - E_{p_\lambda}[\mathbf{f}(x_j, y)])$. This gradient matches model moments of features $\mathbf{f}$ to the empirical moments on labeled data $\mathcal{D}$ and auxiliary moments on unlabeled data $\mathcal{D}'$.

We call the above optimization algorithm *alternating projections* (AP). Since gradients in both projection steps decompose according to the instances, we can perform online optimization using stochastic gradient descent (Bottou, 2004):

1. For $t = 1, \ldots, T$, let $\eta = \frac{1}{t+t_0}$ where $t_0 = 1/\eta_0$, $\eta_0$ the initial learning rate. Let $u$ be the target expectations. Let labeled and unlabeled data set sizes be $m$ and $n - m$ respectively. Let the initial parameters be $\lambda^{(0)}$ and $\mu^{(0)}$.

2. For a new *labeled* instance $x_t$ with label $y_t$, set $\mu^{(t)} = \mu^{(t-1)}$ and $\lambda^{(t)} = \lambda^{(t-1)} + \eta[\mathbf{f}(x_t, y_t) - E_{p_{\lambda^{(t-1)}}}[\mathbf{f}(x_t, y)] - \frac{\alpha\lambda^{(t-1)}}{n}]$.

3. For a new *unlabeled* instance $x_t$, set $\mu^{(t)} = \mu^{(t-1)} + \eta[\frac{u}{(n-m)} - E_{q_{\lambda^{(t-1)},\mu^{(t-1)}}}[\mathbf{f}'(x_t, y)] - \frac{\beta\mu}{(n-m)}]$ and $\lambda^{(t)} = \lambda^{(t-1)} + \eta[E_{q_{\lambda^{(t-1)},\mu^{(t-1)}}}[\mathbf{f}(x_t, y)] - E_{p_{\lambda^{(t-1)}}}[\mathbf{f}(x_t, y)] - \frac{\alpha\lambda^{(t-1)}}{n}]$.

This enables scaling our approach to large data sets. Recent empirical evidence demonstrates the usefulness of this online optimization procedure on real-world tasks of sequence tagging (Liang et al., 2009) and dependency parsing (Ganchev et al., 2009).

## 3 Applications

In this section, we briefly describe the application of AP to several learning problems. First, note that AP can be applied either transductively, where the unlabeled data includes the test data, or inductively, where the model is applied to unseen data. Also, note that constraints may be expressed on a per-instance basis (Graca et al., 2008), or summarizing the entire unlabeled data set (Mann & McCallum, 2008). Per-instance constraints are obtained by defining a feature function $f'$ that only returns a non-zero value for a particular $x$.

*Semi-supervised learning* aims to incorporate available unlabeled data into parameter estimation. We perform semi-supervised learning using AP by first estimating the parameters of the model distribution $p$ using available labeled data. Subsequently, we choose the auxiliary distribution $q$ that is closest to $p$ and additionally satisfies the auxiliary expectation constraints. The predictions of the auxiliary distribution $q$ are then used as soft labels for the unlabeled data. Next, this soft-labeled data is used in addition to the original labeled data to re-estimate the parameters of $p$. The additional constraints on $q$ may come from human prior knowledge about the task, or from the labeled data itself. Importantly, the practitioner has the freedom to specify which expectation constraints they expect to hold on unlabeled data. For example, the practitioner may choose to only apply the most confident and helpful constraints from the labeled data to the unlabeled data.

In *minimally supervised learning* there are no labeled examples but we may encode prior knowledge about the task in the form of a few expectation constraints. These constraints could directly be used in maximum entropy estimation, but the resulting exponential model would only have parameters for the constraints. Often we are aware of other potentially helpful feature functions for the task, but have no prior information about their expectations. For example, in text classification, we know that the words present in the unlabeled data are likely to be relevant for the classification task, but we might only know the precise role of a few words. Using AP, we can learn a feature-rich model $p$ from a few expectation constraints. In general, this model will have better generalization performance than a model that only has parameters for constraint features.

Finally, it is important to note that the auxiliary distribution $q$ and the model distribution $p$ may have different parameterizations. It is exactly this fact that allows us to use a small number of expectation constraints to guide the learning of a feature-rich model (as described above). Another way in which we can leverage this flexibility is to include constraints that look at more output variables than the feature functions in $p$. For example, although $p$ may be a first-order linear chain CRF, $q$ may express *long-range constraints* over the entire sequence. Although this requires approximate inference while learning $q$, the final model $p$ can be efficiently applied at test time.

# 4 Related Work

There has been much recent interest in learning with auxiliary expectation constraints. In this section we describe the relationship between our framework and previously proposed methods.

## 4.1 Constraint-driven Learning

Constraint-driven learning (Chang et al., 2007) (CODL) is an EM-like algorithm that incorporates per-instance constraints into semi-supervised training of structured output models. In the E-step, the inference procedure produces an $n$-best list of outputs ranked according to the sum of the model score and a term that penalizes violated constraints. In the M-step, the $n$-best list is used to re-estimate the model parameters. AP has two important advantages over CODL. First, CODL uses point estimates to re-estimate the model distribution, whereas we maintain a complete distribution over output variables. Second, the constraint parameters in CODL are set manually. In contrast, we ask practitioners to provide expectation values that are implicitly mapped to parameter values. The language of expectations is likely more interpretable to a human than the language of parameter values.

## 4.2 Posterior regularization

Posterior regularization or EM with posterior constraints (EMPC) (Graca et al., 2008) is a modified EM algorithm in which the E-step is replaced by the information projection of the model posterior distribution onto the set of distributions that satisfy auxiliary expectation constraints. Note that replacing the conditional model $p_\lambda(y|x)$ in our objective (Equation (6)) with a joint a model $p_\lambda(y, x)$ yields the objective function optimized by Graca et al. (2008). Recently, this framework has also been applied to discriminative models (Ganchev et al., 2008; Ganchev et al., 2009). We arrive at the same objective function and optimization algorithm from the alternate perspective of generalized maximum entropy (Dudik, 2007), which provides additional justification for the use of EM in a discriminative model.

Although AP and EMPC share the same objective, applications of EMPC have primarily used per-instance equality and inequality constraints on output variables. In this paper, we describe the use of $\mathbb{L}_1$ and $\mathbb{L}_2$ penalties for soft matching of target expectations. We are also interested in the use of per-dataset constraints that additionally consider input variables. Empirical results show that such constraints are very helpful in minimally supervised learning.

Importantly, EMPC has not been compared to related frameworks for learning with unlabeled data and auxiliary expectation constraints such as GE and CODL. This is the first paper to provide such a comparison.

## 4.3 Learning from Measurements

Liang et al. (2009) simultaneously developed a Bayesian decision-theoretic framework for learning an exponential family model using general *measurements* on the unlabeled data. These measurements can be labels, partial labels and other constraints on the model predictions. By using a variational inference algorithm, they arrive at an objective that is similar to the one proposed in this paper. Hence, they provide yet another interpretation of the learning algorithm.

## 4.4 Generalized Expectation Criteria

Generalized expectation (GE) criteria express preferences on the model expectation of some feature function. Let $p_\lambda(y|x)$ be a conditional random field model, $f'$ be a constraint function, $\mathcal{D}'$ be unlabeled data, and $U$ be a score function. A GE criterion is then

$$\mathcal{G}(\lambda) = U\big(\sum_{x \in \mathcal{D}'} E_{p_\lambda(y|x)}[f'(x,y)]\big). \qquad (10)$$

Learning with these criteria proceeds by direct optimization of an objective function that contains labeled data likelihood (if there is labeled data), a prior on parameters, and the sum of multiple GE terms:

$$\mathcal{O}(\lambda) = \mathcal{L}(\lambda) + p(\lambda) + \sum_\mathcal{G} \mathcal{G}(\lambda). \qquad (11)$$

When $U$ is the squared difference from some target expectation $u$, the partial derivative of $\mathcal{G}$ with respect to a model parameter for some feature function $f_i$ is the model expected covariance between the constraint function $f'$ and $f_i$,

$$\frac{\partial}{\partial \lambda_i} \mathcal{G}(\lambda) = 2(u - f'_\lambda)$$
$$\Big(\sum_{x \in \mathcal{D}'} E_{p_\lambda(y|x)}[f'(x,y)f_i(x,y)]$$
$$- E_{p_\lambda(y|x)}[f'(x,y)] E_{p_\lambda(y|x)}[f_i(x,y)]\Big), \qquad (12)$$

where $f'_\lambda = \sum_{j=m+1}^n E_{p_\lambda(y|x_j)}[f'(y, x_j)]$. Therefore, GE optimization can be interpreted as a bootstrapping method that adjusts parameters for model feature functions according to their model predicted similarity with constraint functions. This covariance allows GE to learn from long range interactions in graphical models, for example, with feature functions that never directly overlap but do occur within the same instance.

Unfortunately, computing this covariance is costly, as the first term requires the computation of a marginal distribution over as many possible distinct output variables as $f_i$ and $f'$ consider. For a logistic regression model in which constraint functions $f'$ correspond to model feature functions, GE training is $O(|\mathcal{Y}|)$. However, for a linear chain CRF, feature functions $f'$ that consider a single label require $O(n|\mathcal{Y}|^3)$ time for inference, where $n$ is the length of the sequence. With constraints on transitions, the running time increases to $O(n|\mathcal{Y}|^4)$, which is impractical for real world data sets.

In contrast, AP only takes the time required to perform inference in the corresponding model if constraint functions $f'$ factor similarly to model feature functions $f_i$. Therefore, learning a linear chain CRF with AP takes $O(|\mathcal{Y}|^2)$ time for single label or transition functions $f'$. Although we may lose some power afforded by the covariance-based parameter update for structured output models, we may compensate for this by including more expressive features over the input sequence.

Additionally, online learning is not straightforward with GE training, because expectations over the entire data set are needed before the parameters can be updated for any single instance.

## 5  Comparison with GE for Minimally Supervised Learning

In this section, we present experiments comparing AP and GE training for the minimally supervised scenario in which we have only unlabeled data and auxiliary constraints. We demonstrate that AP provides comparable accuracy to GE while providing computational advantages.

### 5.1  Classification

We first compare AP and GE training of logistic regression models for text classification. Druck et al. (2008) describe GE training of logistic regression models for document classification with expectation constraints that specify affinities between words and labels. For example, we know that the word "puck" is indicative of label hockey for classifying documents about sports. We use the same set of constraints, processing of the data sets, and 10 unlabeled / test splits as Druck et al. (2008).

In these experiments, AP uses auxiliary model feature functions $f'$ that only check for the presence of a particular word, and if it is present, return $1/c$, where $c$ is the count of the word. With this construction, all expectation constraints are equally weighted, and cor-respond to the constraints used by Druck et al. (2008). For AP, we set the regularization parameters $\alpha$ and $\beta$ to 1 and 0.01, respectively, for all experiments.

| data set and constraints | base | GE | AP |
|---|---|---|---|
| 20 newsgroups (infogain) | 0.643 | **0.704** | 0.680 |
| 20 newsgroups (LDA) | 0.585 | **0.667** | 0.643 |
| movie (infogain) | 0.772 | 0.797 | **0.800** |
| movie (LDA) | 0.607 | **0.623** | 0.621 |
| sector top (infogain) | 0.719 | **0.730** | 0.726 |
| sector top (LDA) | 0.544 | 0.596 | **0.625** |
| sraa (infogain) | 0.585 | 0.651 | **0.713** |
| sraa (LDA) | 0.520 | 0.559 | **0.589** |
| webkb (infogain) | 0.745 | **0.774** | 0.772 |
| webkb (LDA) | 0.593 | **0.615** | 0.588 |

Table 1: Comparison of GE and AP training of logistic regression models using "labeled features". Underlined values indicate that the method significantly outperformed all others. AP achieves similar performance to GE, but could additionally be used in an online setting, whereas GE could not.

The macro-F1 (average of F1 for each class label) results are presented in Table 1. We additionally report the results of a baseline that uses GE but removes unconstrained features, labeled *base* in the table. GE and AP obtain comparable macro-F1, with AP performing significantly better in three experiments, and GE performing better in two. Both methods outperform the baseline in all cases with the exception of AP on *webkb (LDA)*. Although the performance is comparable, AP can additionally be used in an online setting.

### 5.2  Sequence Labeling

We next compare estimating linear chain CRF parameters using GE and AP. Mann and McCallum (2008) describe GE training of CRFs given estimates of the expectations of functions $f'$ that consider a single label and the input sequence. We use the *apartments* data set, in which the task is to segment Craigslist apartment classifieds into fields such as features, rent, size, and neighborhood. There are 11 labels in total. We use the same data processing and the same set of constraints as Mann and McCallum (2008).[1] The constraints specify that, for example, if the word at some position is "call", then the label at that position should be contact. In this experiment we use 200 unlabeled examples (the train and dev splits of the *apartments* data) and no labeled examples.

Despite not having the advantage of the covariance term to spread parameter weight (see Section 4.3),

---

[1] See (Mann & McCallum, 2008) for details.

we obtain token accuracy of 61.2% with AP using the same variance settings as in the previous section. If we add an additional transition constraint that specifies that 90% of all transitions should be self-transitions, we obtain an accuracy of 68.1%, better than the 66.6% accuracy obtained by GE (Mann & McCallum, 2008).

Importantly, computing the GE gradient takes $O(n|\mathcal{Y}|^3)$ time for a sequence of length $n$ and constraint functions that consider a single label, whereas AP takes $O(n|\mathcal{Y}|^2)$. Additionally, including the self-transition constraint with GE would require $O(n|\mathcal{Y}|^4)$ time, which is impractical for real data sets with tens of labels. In contrast, expressing the same constraint in AP still takes $O(n|\mathcal{Y}|^2)$ time, a quadratic speed-up.

## 6 Higher-order Constraint Experiments

In this section, we consider the effect of adding higher-order constraints for a sequence segmentation and labeling task. We present empirical results on the Cora reference extraction task (Peng & McCallum, 2004) which consists of 500 labeled citations of computer science research papers. We used a tokenized version of the data set provided by (Chang et al., 2007) with a similar data split of 300/100/100 training, development and test examples. The label set consists of 13 labels: author, editor, title, booktitle, journal, date, volume, pages, publisher, tech, institution, location and note. In addition to the Cora data, we used an unlabeled data set with 1000 citations identical to (Chang et al., 2007).

Our model $p(\mathbf{y}|\mathbf{x})$ is a linear chain conditional random field (CRF) with edges between subsequent labels of the label sequence $\mathbf{y}$. Hence, the features in our model have the functional form $f(y_{t-1}, y_t, \mathbf{x})$. Our CRF model allows us to use rich input features: **(1)** *token-based features* including identity, token prefixes, token suffixes and character n-grams, **(2)** *lexicon features* detect the token is presence of a token in a lexicon of author names, journal names, etc., **(3)** *regular expressions* detect common patterns for years and page numbers, and **(4)** token, lexicon and regex features within a fixed window of the token position.

We compare our method against the constraint-driven learning framework (**CRR07**) (Chang et al., 2007). The constraints used in our experiments are identical to **CRR07** and are shown in Table 2. The auxiliary distribution $q(\mathbf{y}|\mathbf{x})$ has feature functions for each of these constraints. The constraint values $\boldsymbol{u}$ are uniformly set to 0.99 for constraints (3)-(11). For constraint (2) we use a feature that tests whether there is a transition on a non-punctuation character and a constraint value 0.01. For (1) we use the constraint $1.0 \geq E[f(\mathbf{y})]$ where feature $f(\mathbf{y})$ counts the number of field repetitions in a label sequence $\mathbf{y}$.

| 1) Each field is a contiguous sequence of tokens and appears at most once in a citation |
|---|
| 2) Transitions between fields occur on punctuation marks |
| 3) The citation can only start with author or editor |
| 4) The words *pp*, *pages* correspond to page |
| 5) Four digits starting with 20xx and 19xx are date |
| 6) Quotations can appear only in titles |
| 7) The words *note*, *submitted* and *appear* have label note |
| 8) The words *CA*, *Australia* and *NY* are location |
| 9) The words *tech*, *technical* have label tech |
| 10) The words *proc*, *journal*, *proceedings*, *ACM* are either journal or booktitle |
| 11) The words *ed*, *editors* correspond to editor |

Table 2: The list of constraints used in our citation extraction task. The first entry in the table is a higher order constraint that needs to look at the entire label sequence.

Since our auxiliary distribution now uses higher-order constraints we cannot perform exact inference in our model to obtain feature expectations. We instead use approximate inference in our auxiliary model using Gibbs sampling. Additionally, to speed up training, we use stochastic gradient descent (SGD) with an initial learning rate $\eta = \frac{1}{t} = 0.1$ where $t$ is incremented after each example (Bottou, 2004).

We initially trained a CRF model with increasing number of labeled examples $N = 5, 20, 300$ (**Sup** baseline). Next, we train our CRF model either transductively by applying constraints on the test data (**AP**-T) or inductively by applying constraints on the unlabeled data set (**AP**-I). In both cases, we weight the contribution of the unlabeled data with $\gamma = 0.1$ and use regularization constants $\alpha = 1.0$ (Eq. (9)) and $\beta = 1.0$ (Eq. (8)). We ran the alternating projections method for $T = 10$ iterations. The hyper-parameters of our model are chosen by cross-validation.

| N | **Sup** | **AP**-T | **AP**-I | **CRR07** |
|---|---|---|---|---|
| 5 | 0.622 | **0.756** | **0.746** | 0.710 |
| 20 | 0.798 | **0.854** | **0.851** | 0.794 |
| 300 | 0.940 | **0.943** | **0.948** | 0.882 |

Table 3: Comparison of different models against constraint-driven learning for varying number of labeled examples $N$. Results are averaged over 5 runs. The bold results indicate significantly greater performance for transductive and inductive learning.

We measure the performance of our models in terms of the token labeling accuracy on test data. The results of reported in Table 3 are averaged over 5 runs with a different random subset of the data. We used the subsets identical to those used by (Chang et al., 2007). We obtain state-of-the-art results on extraction for both transductive and inductive learning. In both cases, we achieve an absolute reduction in error over **CRR07** of 3% with 5 labeled examples and about 6% with 20 and 300 labeled examples. Additionally, our **AP** models improve over the supervised baseline significantly with an error reduction of 12% with just 5 labeled examples.

## 7 Conclusions and Future Work

We presented a general framework for semi-supervised learning with expectation constraints. The results demonstrate the effectiveness of our method in comparison to previous state-of-the-art constraint-driven learning frameworks. In addition, we show that imposing constraints on long-range dependencies between labels significantly reduces error in a sequence extraction task without increasing computational cost.

In future work, we plan to apply our framework to the task of domain adaptation and graph-based label regularization. In this paper, we only experimented with feature-label constraints and a few higher-order constraints. We plan to explore richer convex constraints and their effect on natural language problems.

## Acknowledgments

We thank Kuzman Ganchev and Ben Taskar for their helpful comments. This work was supported in part by the Center for Intelligent Information Retrieval and in part by The Central Intelligence Agency, the National Security Agency and National Science Foundation under NSF grant #IIS-0326249. Any opinions, findings and conclusions or recommendations expressed in this material are the authors' and do not necessarily reflect those of the sponsor.